\documentclass[11pt]{article}

\usepackage[a4paper, margin=2.5cm]{geometry}
\usepackage{times}
\usepackage{titlesec}
\usepackage{parskip}

\usepackage{amsmath, amssymb,mathtools}

\usepackage{graphicx}
\usepackage{caption}
\usepackage{subcaption} 
\usepackage{float}
\usepackage{booktabs}
\usepackage{multirow}
\usepackage{tabularx}
\usepackage{nicematrix}
\usepackage{blkarray}
\usepackage{array,ltablex,multirow}
\usepackage{longtable}
\usepackage{makecell}
\usepackage{flushend} 

\usepackage{adjustbox}

\usepackage[square,numbers,sort&compress]{natbib}

\usepackage[hidelinks]{hyperref}
\usepackage[capitalize, noabbrev, nameinlink]{cleveref}

\newcommand{\define}[1]{{\bf \boldmath{#1}}}

\title{\bfseries The persistence of painting styles}
\author{Reetikaa Reddy Munnangi\textsuperscript{1} \and Barbara Giunti\textsuperscript{2}\footnotemark
}
\date{}

\begin{document}

\maketitle

\begin{abstract}
Art is a deeply personal and expressive medium, where each artist brings their own style, technique, and cultural background into their work. 
Traditionally, identifying artistic styles has been the job of art historians or critics, relying on visual intuition and experience. 
However, with the advancement of mathematical tools, we can explore art through more structured lenses. 
In this work, we show how persistent homology (PH), a method from topological data analysis, provides objective and interpretable insights on artistic styles. 
We show how PH can, with statistical certainty, differentiate between artists, both from different artistic currents and from the same one, and distinguish images of an artist from an AI-generated image in the artist's style. 
\end{abstract}

\section{Introduction}

Humans have been creating art since the beginning of their time, and each person has their peculiar way of doing it. 
The fact that each artist has their unique ``style'' and that, nevertheless, different artists may fall into the same ``artistic current'', i.e., their ``styles'' are similar, appears evident. 
However, agreeing on an objective, formal definition of either ``artistic style'' or ``artistic current'' seems impossible, since the perception of art is inherently subjective, artists evolve, and the borders between currents and styles blend into one another. 
In this work, we will not attempt to solve this century-long problem since we are not art experts, but we will provide quantitative methods to objectively describe the style of a painter using tools from Persistence Theory (PT). 
\newline
PT is a discipline at the intersection of (algebraic) topology, computer science, and data analysis that extracts information from data that usually cannot be recovered using other data analysis methods. 
In particular, information about connectivity and the presence of ``loops'' or ``cavities'', or, more generally, about the shapes in the data. 
Indeed, topology (together with geometry) is the branch of mathematics that studies shapes and their transformations rigorously. 
PT has been proven to be extremely successful in data analysis (see \cite{donut_database,donut_paper} for over 300 examples), but, to the best of our knowledge, has never been applied to painting styles (see \cref{ssec:related_work} for a literature comparison). 
PT has several advantages over other data analysis methods that are particularly useful in art analysis. 
It is not artist- or current-dependent: it is highly adaptable beyond the works of art analyzed in this article. 
All analyses we performed can be extended to different collections of paintings, simply using the code and instructions we provide at \cite{github_repo}.
It does not need training databases, it is interpretable and provides nuanced descriptions. 
Last but not least, it captures features at all scales: from the minute brush strokes to the largest (color) components, the only limitation being the resolution of the image.
\newline
Persistent Homology (PH) \cite{dey2022,carlsson2009, oudot2015,munch2017} is the PT tool we will be using.
Raw data usually needs to be preprocessed before it can be analyzed. 
Several standard preprocessing techniques require choosing a threshold and fixing it for the remainder of the analysis. 
This choice is typically arbitrary (even if guided by expert intuition), and thus the analysis is subjective and, often, unstable with respect to different choices. 
PT overcomes this limitation by analyzing all thresholds at once. 
Therefore, the output is no longer a single value, but rather the evolution of the value across the range of thresholds. 
When using PH, this value is the \emph{($n$)-homology} of the data. 
For $n=0,1$, the $n$-homology is the number of connected components, resp. loops, in the data (see \cref{sec:tools} more details). 
Therefore, the $n$-PH of an object is the evolution (often referred to as \emph{($n$)-barcode} or \emph{($n$)-persistence diagram}) of the $n$-homology of the object across the range of one threshold.
\newline
In this work, we use PH to answer the following questions: 
\begin{itemize}
    \item[\hypertarget{(i)}{(i)}] How can we distinguish artists from different currents? 
    \item[\hypertarget{(ii)}{(ii)}] How can we distinguish artists from the same current?
    \item[\hypertarget{(iii)}{(iii)}] Is it possible to distinguish between the paintings of an artist and an AI-generated image in their style?
\end{itemize}
In detail, we collected 10 paintings each for several different artists, both Eastern and Western, from WikiArt. 
We turned each of them into five channels: their three RGB channels, and the grayscaled and the edge-detected \cite{edge} version of the painting. 
Then we computed, for each channel, two topological features, namely the $0$-PH and the $1$-PH. 
We thus calculated the distances (bottleneck and $1$-Wasserstein \cite{CEH07,CohenSteiner2010}, see \cref{sec:tools} for details) between all $n$-signatures for each channel, for $n=0,1$. 
We then performed several permutation tests \cite{permtest_Fisher,permtest_pitman}, first to classify ten artists from different artistic currents, then to distinguish five artists from the same artistic current (Impressionism and Academic Realism), and, finally, to tell apart images from chosen artists (Klimt and Raja) from an AI-generated image in their styles.
The results were impressive: several topological signatures (possibly different ones for each case) are able, with statistical certainty, to differentiate all cases.
Moreover, both the significant and, surprisingly, the non-significant outputs give us quantitative insights on the style of the painters: for example, from a topological point of view, Siddharth Shingade applies his red and green nuances uniquely, but his blue nuances are indistinguishable from how other artists use them (see \cref{tab:nocurrent_wasserstein} and \cref{sec:experiments}).
The Python code we used for the experiments is openly available at \cite{github_repo}. 
While readers are encouraged to use it on their own set of paintings, the collections we used for our experiments are available at \cite{our_zenodo} to ensure the reproducibility of our findings.
\newline
This work opens an exciting avenue for research, combining applied topology and art, providing quantitative insights from persistence theory to create a strong, formal base for artistic style analysis.

\subsection{Related Work}\label{ssec:related_work}

While, from what we could ascertain, this is the first article on PT and art, there has been previous work about different mathematical tools and art. 
The works closest to ours are \cite{fractal_betti_for_complexity,Dmitruk2026ArtsHiddenTopology}; the former uses Betti numbers, a topological invariant, and fractal dimension to analyze the style of one specific painter, Yayoi Kusama, in comparison with Jackson Pollock. 
The latter appeared some months after the first preprint version of our work was made public, and uses persistent homology applied to paintings but it focuses on the relation between artworks and audience, instead of on the insights that PT can give us on arts and styles.
\cite{multifractal2} and its follow-up \cite{multifractal1} also use fractals, in this case to study (grayscaled) expressionist paintings, proving that (multi)fractal analysis is able to distinguish between different expressionist painters, even if it is not able to distinguish artists from different currents. 
Topology in art is also discussed in \cite{topology_for_art_shapes}, where the author proposes a topological set-up tailored to design.
Other mathematical tools, such as entropy, dimensionality reduction, and stochastic evaluations are applied in \cite{art_entropy, escher_dimred_grayscale} and \cite{stochastic_evaluation}, respectively, to encode the complexity of paintings, but they are either restricted to specific artists \cite{escher_dimred_grayscale,stochastic_evaluation} or limited to quantify how ``complex'' a painting is, rather than providing insights on the artistic style. 
\newline
In a different line of work, \cite{neural_arthistory,neural_formalproxy,art_clustering} use neural networks to study artistic styles and currents, but incur the standard limitation of machine learning: huge training databases are needed; when novel images are added, the machine often needs to be retrained; and it is hard to interpret the results since the analysis is performed by a black box.
\newline
Lastly, there has been some work in the opposite direction: using art to retrieve or explain mathematical concepts \cite{art_for_math1,art_for_math2}. 
This line of work is orthogonal to our scope.

\section{Mathematical tools}\label{sec:tools}

We will present a concise, intuitive definition of (persistent) homology, referring the reader to \cite{Hatcher2002,munch2017,oudot2015,dey2022} for more details.
The $n$-homology (over a field) of a (suitable) space $K$ associates to $K$ a vector space $H_n(K)$ whose generators are the $n$-dimensional loops in the space. 
Intuitively, a $0$-dimensional loop is the ``blank space'' between connected components (or clusters), a $1$-dimensional loop is a loop in the common interpretation (i.e., a circle which may have been stretched, like a rubber band), and a $2$-dimensional loop is a ``cavity'', like the inside of a basketball or a football. 
A \define{filtration} $\mathcal{F}$ is a nested sequence of suitable spaces $K_i$: $\mathcal{F}\coloneqq K_0\subseteq \dots \subseteq K_n$. 
In practice, a filtration is made of the data preprocessed with increasing thresholds\footnote{or decreasing, depending on what the threshold represents.}.
The \define{($n$)-PH} of a filtration $\mathcal{F}$ is the $n$-homology of $\mathcal{F}$.
In other words, it is the combined information of the $n$-homology of each space $K_i$ together with its evolution along the sequence of inclusions. 
Therefore, it is often depicted as a \define{barcode}, i.e., a collection of \define{intervals} $[a,b)$, each representing the lifespan of a class in the $n$-homology vector space: the time $a$ represents the threshold value at which the given $n$-homology class was born, i.e., appeared in the thresholded data, and $b$ the value at which that class disappears, i.e., it is no longer in the thresholded data. 
\newline
In this article, the ``suitable'' spaces are \emph{cubical complexes}.
A \define{cubical complex} is a collection closed under subsets of elements of the form $\prod_i J_i$, where $J_i$ is either $\{0\}$, $\{1\}$, or the interval $[0,1]\subseteq \mathbb{R}$. 
A \define{vertex} is just $J_i$, where $J_i$ is one of the singletons, and \define{edge} is just $J_i=[0,1]$, a \define{square} is $\prod_{i=0}^{1}J_i$ where $J_i$ are the interval $[0,1]$, and so forth.
\newline
The two main distances used to compare barcodes are the bottleneck distance \cite{CEH07} and the $p$-Wasserstein distance \cite{CohenSteiner2010}. 
In this paper, we will fix $p=1$ since it is enough for our goals and is the most discriminative.
In both cases, the idea is to match the intervals in the two $n$-PHs by minimizing their distances, and take the distance between any unmatched interval $[a,b)$ and the artificially added interval $[\dfrac{a+b}{2},\dfrac{a+b}{2})$. 
The bottleneck distance is then given by taking the maximal $L_\infty$ distance between matched intervals or between an unmatched interval and its artificially added interval, whichever is greater. 
The $1$-Wasserstein is computed by adding all the $L_1$ distances between matched intervals, and between unmatched intervals and the artificially added intervals. 
\newline
The last mathematical tool we will employ comes from statistics, not applied topology, and we will use it to certify that the bottleneck and $1$-Wasserstein distances that we compute between the PH of paintings are different in a statistically significant way. 
A \define{permutation test} is an exact statistical hypothesis test that assumes that two given samples come from the same distribution \cite{permtest_Fisher,permtest_pitman}. 
Let samples $A$ and $B$ have cardinality $n$ and $m$, respectively, and consider a \define{test statistic} of choice, i.e., a (real) value that can be assigned to the samples. 
Classic examples of test statistics include the mean or the standard deviation, if the samples contain random variables. 
In our case, the samples will be the $n$-PHs of a channel of two sets of paintings, for example from two different artists.
Our test statistic will be the average bottleneck or $1$-Wasserstein distances between the samples, i.e., between the $n$-PH of a painting from one set and the one from the other set.
One first computes the test statistic for the samples $A$ and $B$. 
Then one considers all or, more often since there could be billions of possible permutations, a considerable number of subdivisions of $A\cup B$ into a sample of size $n$ and a sample of size $m$, and records the test statistics for the new samples in each of these subdivisions.
Finally, one plots all the collected test statistics, highlighting the one from the initial samples.
If more than 97.5\% of the test statistics obtained by reshuffling the samples are either below or above the original one, then one can conclude that having obtained that value is extremely unlikely, and therefore the two samples are statistically distinct. 
An important aspect of a permutation test is the number of permutations performed: ideally, one would perform all possible permutations, but this quickly becomes unfeasible if the samples are ``large'', i.e., they contain more than 10 elements each. 
Therefore, the question is: what is the lowest number of permutations needed to make the statistic meaningful, if, for example, one is interested in the 5\% percentile (as we are). 
In \cite{Marozzi_2004}, the author discusses both theoretical and empirical estimates, concluding that 5'000 permutations provide a meaningful answer. 
To be safe, we performed 10'000 permutations, i.e., the suggested number to have meaningful statistics at the 1\% percentile.

\section{Data}

All images of the paintings have been taken from WikiArt \cite{wikiart}. 
We chose a variety of painters spanning different centuries, from Western countries and India, with the goal of showcasing the versatility and capabilities of PT.
We remark that our analysis is not restricted to an artistic current or style, nor to a geographical location or temporal frame. 
The choice of painters and paintings was made by the authors based on their taste.
\newline
For the first set of experiments, we chose the painters: 
Robert Delaunay, Manishi Dey, Artemisia Gentileschi, Francisco Goya, Gustav Klimt, Mithila Art, Raja Ravi Varma, Pierre-Auguste Renoir, Siddharth Shingade, and Vincent van Gogh.
We then selected ten paintings for each of them, which can be found at \cite{our_zenodo}. 
\newline
For our second set of experiments, each author chose an artistic current of her liking: Academic Realism and Impressionism. 
We then considered five artists for each current, and again ten paintings for each painter. 
In detail, the five Impressionist painters we selected are 
Édouard Manet, Claude Monet, Camille Pissarro, Pierre-Auguste Renoir, and Vincent van Gogh. 
The five Academic Realism painters we picked are 
Sawlaram Lakshman Haldankar, Hemendranath Mazumdar, M. F. Pithawala, Raja Ravi Varma, and Jamini Roy. 
These images are available at \cite{our_zenodo} too.
\newline
For our last set of experiments, we first took the ten paintings of Klimt and then the ten of Raja we selected previously, and asked ChatGPT to use them to create first an image ``in Klimt style'' and then ``in Raja style''. 
Specifically, we used the following prompt (after having uploaded the said ten images): ``Consider the ten images I uploaded by Gustav Klimt. Can you generate a new image with similar subjects and style? It has to look as much as possible as if it were a Klimt painting, but it should not be one of his own from the internet.''
Both AI-generated images are available at \cite{our_zenodo}.

\begin{figure*}[!h]
\begin{subfigure}[h]{1\linewidth}
\includegraphics[scale=.43]{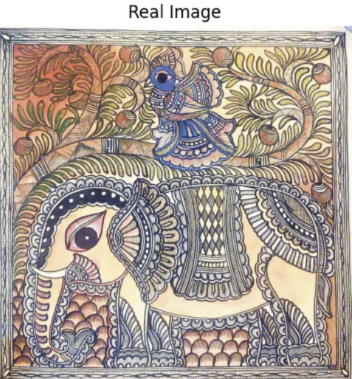}
\label{subfig_originalnochannel}
\end{subfigure}

\begin{subfigure}[t]{.19\linewidth}
\centering
\includegraphics[scale=.52]{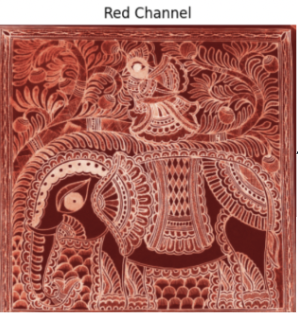}
\label{subfig_redchannel}
\end{subfigure}
\begin{subfigure}[t]{.195\linewidth}
\includegraphics[scale=.51]{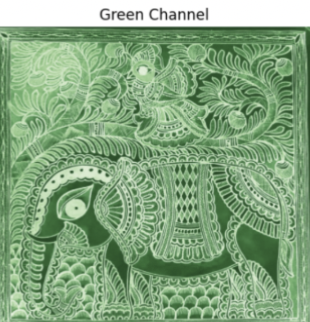}
\label{subfig_greenchannel}
\end{subfigure}
\begin{subfigure}[t]{.195\linewidth}
\includegraphics[scale=.51]{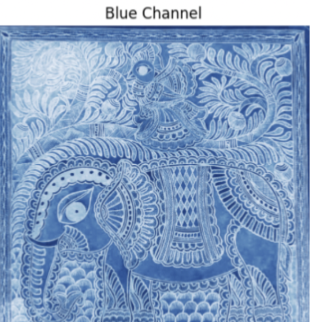}
\label{subfig_bluechannel}
\end{subfigure}
\begin{subfigure}[t]{.19\linewidth}
\includegraphics[scale=.48]{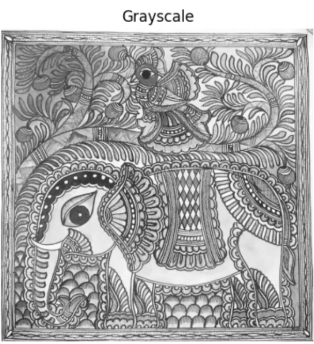}
\label{subfig_graychannel}
\end{subfigure}
\begin{subfigure}[t]{.19\linewidth}
\includegraphics[scale=.515]{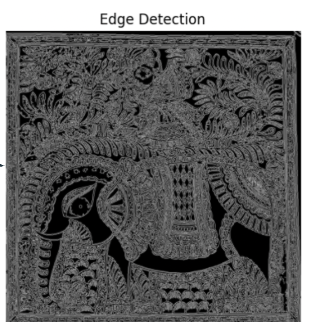}
\label{subfig_edgechannel}
\end{subfigure}
\caption{Original painting from the Mithila Art (top), and its five channels, in order: red, green, blue, grayscale, and edge detection.}
\label{fig:channels}
\end{figure*}

\section{From raw data to filtrations}\label{sec:raw2filt}

\begin{figure}[!h]
\centering
\begin{subfigure}[t]{.48\linewidth}
\includegraphics[scale=.42]{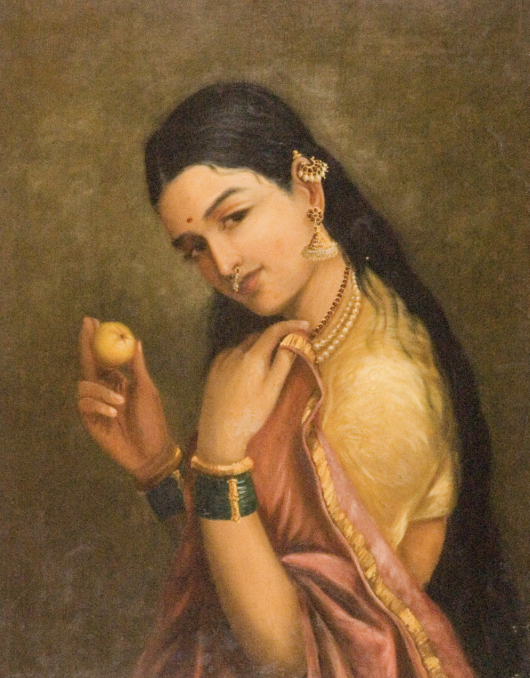}
\label{subfig_original_apple}
\end{subfigure}
\begin{subfigure}[t]{.48\linewidth}
\includegraphics[scale=.254]{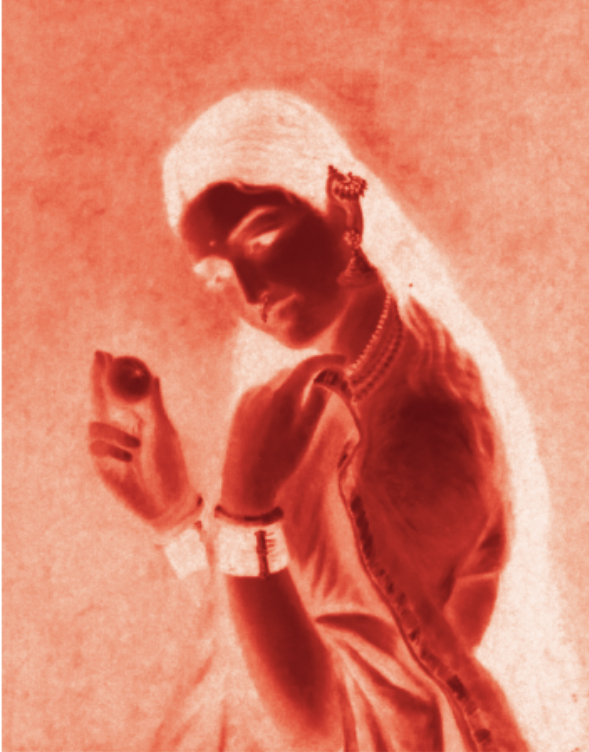}
\label{subfig_red_apple}
\end{subfigure}
\caption{Original painting by Raja Ravi Varma (left) and its red channel (right).}
\label{fig_apple_w_red}
\end{figure}

\begin{figure}[!h]
\begin{subfigure}[t]{.3\linewidth}
\centering
\includegraphics[scale=.1]{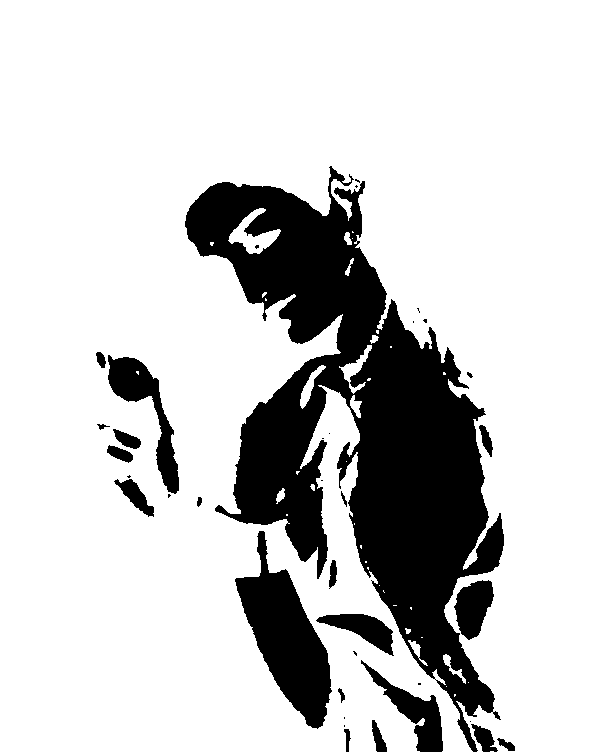}
\caption{$t=100$.}
\label{subfig_bin_t100}
\end{subfigure}
\begin{subfigure}[t]{.3\linewidth}
\hspace{1cm}
\includegraphics[scale=.1]{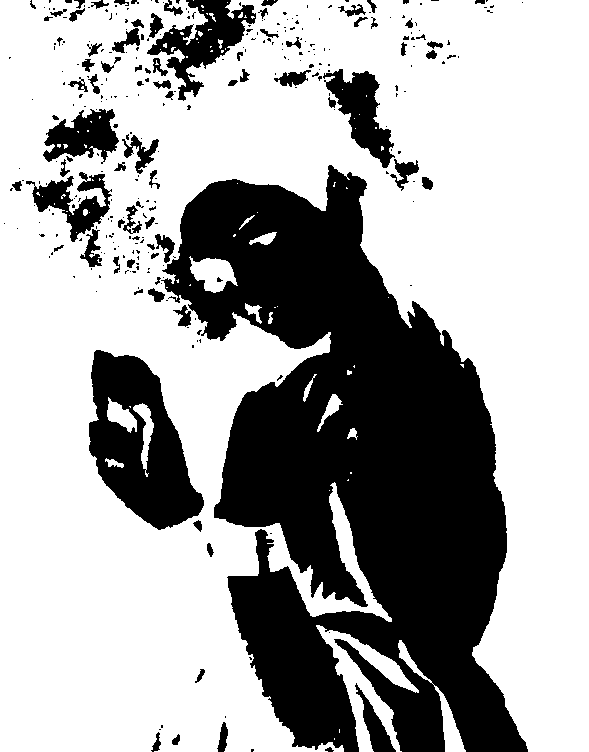}
\caption{$t=150$.}
\label{subfig_bin_t150}
\end{subfigure}
\begin{subfigure}[t]{.3\linewidth}
\hspace{1cm}
\includegraphics[scale=.1]{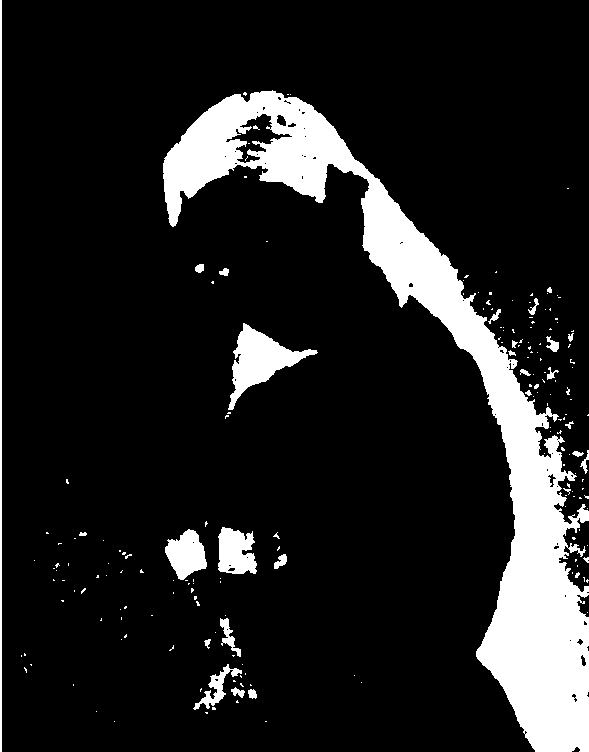}
\caption{$t=200$.}
\label{subfig_bin_t200}
\end{subfigure}
\caption{Examples of binarizations for different thresholds of the red channel in \cref{fig_apple_w_red}.}
\label{fig_bin}
\end{figure}

As explained in \cref{sec:tools}, to compute PH, we need a filtration. 
Our data is a collection of paintings, stored in the machine as RGB images. 
Therefore, we first transform each image into five channels: one each for the RGB colors, the grayscaled version of the image to account for the intensity of the colors together, and a black and white copy of the image with the contours highlighted, also called \define{edge detection} \cite{edge}, see \cref{fig:channels} for an example of all channels, and \cref{fig_apple_w_red} for an example of the red channel.
\newline
Each channel is now stored as a matrix, whose value at index $ij$ is the intensity of the corresponding pixel in the channel, on an integer scale from $0$ to $255$. 
We then obtain a filtration by constructing a cubical complex for each $t=0,\dots,255$.
\newline
To construct a cubical complex from a fixed $t$, we used the \texttt{cubical complex} package in \cite{gudhi}. 
In more detail, we binarize a channel by setting one of its pixels to black if the pixel's value is $\leq t$. 
Each black pixel is a vertex of the cubical complex. 
Adjacent black pixels have an edge between them, and a square is inserted if four black pixels are adjacent in one corner. 
As \cref{fig_bin} shows, the number of black vertices increases with $t$, so each cubical complex includes into the next one, thus forming a filtration.

\section{Experiments and results}\label{sec:experiments}

We performed three sets of experiments, one each for the \hyperlink{(i)}{(i)}--\hyperlink{(iii)}{(iii)} questions we posed in the introduction. 
The setting is the same for each set of experiments: there are two samples ( = collections of channels), $A$ and $B$, which we wish to distinguish with statistical certainty. 
We do so using a permutation test where the test statistic is the average of all the bottleneck (resp. $1$-Wasserstein) distances between the $n$-PH of an element in $A$ and an element in $B$, for $n=0,1$.
\textbf{Disclaimer:} We rounded the $p$-values to three digits, and, as a result, we obtained some 1s and 0s; however, none of the computed $p$-values was either 1 or 0. 
We also output whether the initial sample had the minimum or maximum average distance. 
This information is particularly useful in the case of comparison with AI, since there we will not have enough data to perform a proper permutation test, but the fact that the average distance is either the minimum or the maximum means that the corresponding painting clearly stood out when compared with all of the others.

\begin{table}[h]
\caption{P-values corresponding to the artist's paintings for the bottleneck distances for $0$-PH (left) and $1$-PH (right) across the five channels. Bold values indicate significant results; the symbol $^\ast$ (resp. $_\ast$) means the initial sample had the maximum (resp. minimum) average distance.}
\begin{adjustbox}{width=.49\linewidth}
\begin{NiceTabular}{l*{5}{c}}
\toprule
\multirow{2}{*}{Artist} &
\multicolumn{5}{c}{$0$-PH Bottleneck} \\
\cmidrule(lr){2-6}
& Red
& Green
& Blue
& Gray
& Edge \\
\midrule
Delaunay & 0.96 & 0.94 & 0.76 & 0.78 & \textbf{0.98} \\
\rowcolor{gray!10}Dey & \textbf{0.01} & 0.34 & \textbf{0.00$_\ast$} & 0.64 & 0.18 \\
Gentileschi & \textbf{0.00$_\ast$} & \textbf{0.01} & \textbf{0.00$_\ast$} & 0.05 & 0.45 \\
\rowcolor{gray!10}Goya & \textbf{0.00$_\ast$} & 0.26 & \textbf{0.02} & 0.25 & \textbf{1.00} \\
Klimt & 0.11 & 0.56 & \textbf{0.98} & 0.87 & 0.13 \\
\rowcolor{gray!10}Mithila & 0.87 &\textbf{1.00}  & \textbf{1.00$^\ast$} & \textbf{1.00} & 0.94 \\
Raja & \textbf{0.00$_\ast$} & \textbf{0.00$_\ast$} & \textbf{0.00$_\ast$} & \textbf{0.00$_\ast$} & \textbf{0.00} \\
\rowcolor{gray!10}Renoir & \textbf{0.00$_\ast$} & \textbf{0.00$_\ast$} & \textbf{0.00$_\ast$} & \textbf{0.00$_\ast$} & \textbf{0.00$_\ast$} \\
Shingade & \textbf{0.00$_\ast$} & \textbf{0.00$_\ast$} & \textbf{0.00$_\ast$} & \textbf{0.00$_\ast$} & \textbf{0.02} \\
\rowcolor{gray!10}van Gogh & \textbf{0.00$_\ast$} & \textbf{0.00$_\ast$} & \textbf{0.00$_\ast$} & \textbf{0.00$_\ast$} & \textbf{0.00$_\ast$} \\
\bottomrule
\end{NiceTabular}
\end{adjustbox}
\begin{adjustbox}{width=.49\linewidth}
\begin{NiceTabular}{l*{5}{c}}
\toprule
\multirow{2}{*}{Artist} &
\multicolumn{5}{c}{$1$-PH Bottleneck} \\
\cmidrule(lr){2-6}
& Red
& Green
& Blue
& Gray
& Edge \\
\midrule
Delaunay & 0.75 & 0.67 & 0.67 & 0.47 & 0.82 \\
\rowcolor{gray!10}Dey & 0.32 & 0.47 & 0.91 & 0.61 & \textbf{0.00$_\ast$} \\
Gentileschi & 0.32 & 0.79 & 0.61 & 0.86 & \textbf{0.00$_\ast$} \\
\rowcolor{gray!10}Goya & \textbf{0.00$_\ast$} & \textbf{0.00$_\ast$} & 0.39 & \textbf{0.00$_\ast$} & \textbf{0.00$_\ast$} \\
Klimt & \textbf{0.00$_\ast$} & \textbf{0.00$_\ast$} & 0.26 & \textbf{0.00$_\ast$} & \textbf{0.00$_\ast$} \\
\rowcolor{gray!10}Mithila & 0.37 & 0.42 & \textbf{1.00} & 0.25 & \textbf{0.00$_\ast$} \\
Raja & \textbf{0.00$_\ast$} & \textbf{0.00$_\ast$} & 0.04 & \textbf{0.00$_\ast$} & \textbf{0.00$_\ast$} \\
\rowcolor{gray!10}Renoir & \textbf{0.00$_\ast$} & \textbf{0.00$_\ast$} & \textbf{0.00$_\ast$} & \textbf{0.00$_\ast$} & \textbf{0.00$_\ast$} \\
Shingade & \textbf{0.00$_\ast$} & \textbf{0.00$_\ast$} & 0.30 & \textbf{0.00$_\ast$} & \textbf{0.00$_\ast$} \\
\rowcolor{gray!10}van Gogh & \textbf{0.00$_\ast$} & \textbf{0.00$_\ast$} & \textbf{0.00$_\ast$} & \textbf{0.00$_\ast$} & \textbf{0.00$_\ast$} \\
\bottomrule
\end{NiceTabular}
\end{adjustbox}
\label{tab:nocurrent_bottleneck}
\end{table}
In the first set of experiments, we have 100 paintings from 10 artists. 
To prove that they can be differentiated with topological signatures, we perform 10 permutation tests per channel per $n$-signature, with $n=0,1$, each with the sample $A$ with the channel versions of 10 paintings from one fixed artist, and the sample $B$ with the same channel for the remaining 90 paintings. 
Each of these computations was repeated for the five channels, for bottleneck and $1$-Wasserstein, and for $0$- and $1$-PH. 
In each permutation test, we performed 10'000 permutations, i.e., the double of the suggested amount to ensure significant results \cite{Marozzi_2004}.
The resulting $p$-values for the bottleneck distance, respectively for $0$- and $1$-PH, are presented in \cref{tab:nocurrent_bottleneck}, while the $p$-values for the $1$-Wasserstein distance, again divided into $0$- and $1$-PH, are reported in \cref{tab:nocurrent_wasserstein}.
\newline
The results were stunning: almost every artist can be classified by several channels, either in $0$- or $1$-PH, either with the bottleneck or the $1$-Wasserstein distance. 
The only surprising exception is Robert Delaunay: only the bottleneck distance for the edge channel in $0$-PH has a significant $p$-value and can distinguish Delaunay from all other painters. 
This result is particularly interesting since Delaunay is the only abstract painter in this set of experiments, and, therefore, he is very easily singled out by humans.
Nonetheless, the PHs of his paintings are apparently very similar to those of many other paintings, suggesting that topology is capturing some hidden property of the artistic styles and is thus a useful asset, bringing novel information to the study of art.
For all other painters, several topological features can classify them accurately: in \cref{tab:nocurrent_bottleneck}, only 37 $p$-values are not significant, out of the 100 computed. 
Interestingly, the $1$-Wasserstein, which is a more refined distance as it considers all features and not only the longest one, is slightly worse for classification, with 42 non-significant $p$-values in \cref{tab:nocurrent_wasserstein} out of 100 total. 
For both the bottleneck and the $ 1$-Wasserstein distance, the $1$-PH for the edge channel is the most discriminative feature, and the $0$-PH for the blue channel is the second-best.
\newline
Topologically speaking, Renoir and van Gogh each have their own very peculiar styles, as shown by the significance of all their $p$-values for every distance and channel.
In general, the $1$-Wasserstein struggles to distinguish Dey, Gentileschi, and Goya, but the Mithila Art has almost always the maximal average $1$-Wasserstein distance (contrary to what happens with the bottleneck distance), which means that, when considering all intervals in the topological signatures, Mithila Art has a unique distribution of $1$-dimensional loops and of components.
Opposite is the behavior of Raja's paintings: their $p$-values, for all channels and distances, are very low. 
This difference in values has a meaning: when the $p$-values are very close to $1$, it means that, on average, an artist is much farther away from all the other painters, and when the $p$-values approach $0$ it means that they are very close to all of the other painters. 
This also gives us insights about their styles: the artists with very high $p$-values across the board have a very distinctive style, in the sense that their contours, way of clustering colors, and circular structures are very different. 
The painters with low $p$-values across the tables may have a less distinctive patching of colors, or choose more common contours, and therefore be ``close'' to several other artists, and yet their style is no less unique, since these $p$-values are also statistically significant. 
We stress that these descriptions are expressed in mathematical more than artistic terms:the insights of art critics will be needed to interpret our description in artistic terms.
\begin{table}[h]
\caption{P-values corresponding to the artist's paintings for the $1$-Wasserstein distances across for $0$-PH (left) and $1$-PH (right) across the five channels. Bold values indicate \textit{not significant} results; the symbol $^\ast$ (resp. $_\ast$) means the initial sample had the maximum (resp. minimum) average distance.}
\begin{adjustbox}{width=.49\linewidth}
\begin{NiceTabular}{l*{5}{c}}
\toprule
\multirow{2}{*}{Artist} &
\multicolumn{5}{c}{$0$-PH $1$-Wasserstein} \\
\cmidrule(lr){2-6}
 & Red 
 & Green 
 & Blue
 & Gray
 & Edge \\
\midrule
Delaunay & 0.47 & 0.23 & 0.21 & 0.05 & 0.50 \\
\rowcolor{gray!10}Dey & 0.68 & 0.83 & 0.91 & 0.83 & 0.70 \\
Gentileschi & 0.41 & 0.82 & 0.53 & 0.86 & 0.47 \\
\rowcolor{gray!10}Goya & \textbf{0.02} & 0.45 & 0.58 & 0.52 & 0.68 \\
Klimt & \textbf{1.00} & \textbf{1.00} & \textbf{1.00} & \textbf{1.00} & 0.67 \\
\rowcolor{gray!10}Mithila & \textbf{1.00$^\ast$} & \textbf{1.00$^\ast$} & \textbf{1.00$^\ast$} & \textbf{1.00} & \textbf{1.00$^\ast$} \\
Raja & \textbf{0.00$_\ast$} & \textbf{0.00$_\ast$} & \textbf{0.00} & \textbf{0.00$_\ast$} & 0.66 \\
\rowcolor{gray!10}Renoir & \textbf{0.98} & \textbf{0.98} & \textbf{1.00} & \textbf{0.98} & \textbf{1.00$^\ast$} \\
Shingade & \textbf{1.00} & \textbf{0.98} & \textbf{1.00} & \textbf{0.98} & 1.00 \\
\rowcolor{gray!10}van Gogh & \textbf{1.00} & \textbf{1.00} & \textbf{1.00$^\ast$} & \textbf{1.00} & \textbf{1.00$^\ast$} \\
\bottomrule
\end{NiceTabular}
\end{adjustbox}
\begin{adjustbox}{width=.49\linewidth}
\begin{NiceTabular}{l*{5}{c}}
\toprule
\multirow{2}{*}{Artist} &
\multicolumn{5}{c}{$1$-PH $1$-Wasserstein} \\
\cmidrule(lr){2-6}
 & Red
 & Green
 & Blue 
 & Gray
 & Edge \\
\midrule
Delaunay & 0.31 & 0.23 & 0.36 & 0.15 & 0.81 \\
\rowcolor{gray!10}Dey & 0.90 & 0.89 & 0.82 & 0.88 & \textbf{0.00$_\ast$} \\
Gentileschi & 0.75 & 0.85 & 0.56 & 0.87 & \textbf{0.00$_\ast$} \\
\rowcolor{gray!10}Goya & 0.44 & 0.70 & 0.64 & 0.73 & \textbf{0.00$_\ast$} \\
Klimt & \textbf{0.99} & \textbf{0.98} & 0.97 & 0.97 & \textbf{0.99} \\
\rowcolor{gray!10}Mithila & \textbf{1.00$^\ast$} & \textbf{1.00$^\ast$} & \textbf{1.00$^\ast$} & \textbf{1.00$^\ast$} & \textbf{1.00$^\ast$} \\
Raja & \textbf{0.00$_\ast$} & \textbf{0.00$_\ast$} & \textbf{0.00$_\ast$} & \textbf{0.00$_\ast$} & \textbf{0.00$_\ast$} \\
\rowcolor{gray!10}Renoir & \textbf{1.00$^\ast$} & \textbf{1.00} & \textbf{1.00} & \textbf{1.00$^\ast$} & \textbf{0.02} \\
Shingade & \textbf{0.98} & \textbf{0.98} & \textbf{1.00} & 0.95 & \textbf{0.00$_\ast$} \\
\rowcolor{gray!10}van Gogh & \textbf{1.00$^\ast$} & \textbf{1.00} & \textbf{1.00} & \textbf{1.00} & \textbf{0.01} \\
\bottomrule
\end{NiceTabular}
\end{adjustbox}
\label{tab:nocurrent_wasserstein}
\end{table}

\begin{table}[!h]
\caption{The $p$-values for the Impressionists using bottleneck distances for all channels on $0$-PH (top) and $1$-PH (bottom). Significant $p$-values are in bold; the symbol $^\ast$ (resp. $_\ast$) means the initial sample had the max (resp. min) average distance.}
\begin{adjustbox}{width=.69\linewidth,center}
\begin{NiceTabular}{l*{5}{c}}
\toprule
\multirow{2}{*}{Artist} &
\multicolumn{5}{c}{$0$-PH Bottleneck} \\
\cmidrule(lr){2-6}
 & Red 
 & Green 
 & Blue
 & Gray
 & Edge \\
\midrule
\'Eduard Manet & 0.75 & 0.68 & 0.77 & 0.62 & 0.91 \\
\rowcolor{gray!10}Claude Monet & 0.26 & 0.97 & \textbf{1.00} & 0.97 & \textbf{0.00} \\
Camille Pissarro & \textbf{0.00$_\ast$} & 0.07 & 0.63 & 0.13 & 0.35 \\
\rowcolor{gray!10}Pierre-Auguste Renoir & \textbf{0.00$_\ast$} & \textbf{0.00} & \textbf{0.00} & \textbf{0.00} & 0.12 \\
Vincent van Gogh & \textbf{0.00$_\ast$} & \textbf{0.00$_\ast$} & 0.50 & \textbf{0.00$_\ast$} & \textbf{0.00$_\ast$} \\
\hline
\end{NiceTabular}
\end{adjustbox}
\begin{adjustbox}{width=.69\linewidth,center}
\begin{NiceTabular}{l*{5}{c}}
\toprule
\multirow{2}{*}{Artist} &
\multicolumn{5}{c}{$1$-PH Bottleneck} \\
\cmidrule(lr){2-6}
 & Red 
 & Green 
 & Blue
 & Gray
 & Edge \\
\midrule
\'Eduard Manet & 0.96 & 0.97 & \textbf{0.99} & 0.97 & 0.88 \\
\rowcolor{gray!10}Claude Monet & 0.33 & 0.22 & 0.38 & 0.41 & \textbf{0.00} \\
Camille Pissarro & \textbf{0.00$_\ast$} & \textbf{0.00} & \textbf{0.00$_\ast$} & \textbf{0.00} & \textbf{0.00} \\
\rowcolor{gray!10}Pierre-Auguste Renoir & \textbf{0.02} & \textbf{0.00} & \textbf{0.00} & \textbf{0.00} & \textbf{0.00$_\ast$} \\
Vincent van Gogh & \textbf{0.00$_\ast$} & \textbf{0.00$_\ast$} & \textbf{0.00$_\ast$} & \textbf{0.00$_\ast$} & \textbf{0.00$_\ast$} \\
\hline
\end{NiceTabular}
\end{adjustbox}
\label{tab:impressionists_bottleneck}
\end{table}
The second set of experiments, whose goal is to answer question \hyperlink{(ii)}{(ii)}, is twofold.
First, we studied five Impressionist artists, and then we analyzed five Academic Realism painters. 
In both cases, similarly to what we did in the first set of experiments, we performed five rounds of permutation tests, each with the 10 paintings of one fixed artist in the sample $A$ and the 40 remaining paintings in the sample $B$. 
Also in this case, we performed 10'000 permutations in each permutation test. 
As shown in \cite{Marozzi_2004}, this number of permutations provides a meaningful statistical $p$-value.
To be thorough, instead of restricting to the most promising channels obtained from the first experiments, we repeated each of these tests for all five of them. 
The resulting $p$-values can be seen in \cref{tab:impressionists_bottleneck} and \cref{tab:impressionists_wasserstein} for the Impressionists, and in \cref{tab:ac_bottleneck} and \cref{tab:ac_wasserstein} for the Academic Realists.
\begin{table}[h]
\caption{The $p$-values for the Impressionists using $1$-Wasserstein distances for all channels on $0$-PH (top) and $1$-PH (bottom). Significant $p$-values are in bold; the symbol $^\ast$ (resp. $_\ast$) means the initial sample had the max (resp. min) average distance.}
\begin{adjustbox}{width=.69\linewidth,center}
\begin{NiceTabular}{l*{5}{c}}
\toprule
\multirow{2}{*}{Artist} &
\multicolumn{5}{c}{$0$-PH $1$-Wasserstein} \\
\cmidrule(lr){2-6}
 & Red 
 & Green 
 & Blue
 & Gray
 & Edge \\
\midrule
\'Eduard Manet & 0.50 & 0.79 & 0.54 & 0.73 & 0.87 \\
\rowcolor{gray!10}Claude Monet & 0.42 & 0.30 & 0.92 & 0.46 & 0.03 \\
Camille Pissarro & 0.94 & 0.96 & \textbf{1.00} & 0.96 & 0.19 \\
\rowcolor{gray!10}Pierre-Auguste Renoir & \textbf{0.99} & 0.97 & \textbf{1.00} & 0.97 & \textbf{1.00} \\
Vincent van Gogh & \textbf{1.00$^\ast$} & \textbf{1.00$^\ast$} & \textbf{1.00$^\ast$} & \textbf{1.00$^\ast$} & \textbf{1.00$^\ast$} \\
\hline
\end{NiceTabular}
\end{adjustbox}
\begin{adjustbox}{width=.69\linewidth,center}
\begin{NiceTabular}{l*{5}{c}}
\toprule
\multirow{2}{*}{Artist} &
\multicolumn{5}{c}{$1$-PH $1$-Wasserstein} \\
\cmidrule(lr){2-6}
 & Red 
 & Green 
 & Blue
 & Gray
 & Edge \\
\midrule
\'Eduard Manet & 0.47 & 0.79 & 0.62 & 0.73 & 0.90 \\
\rowcolor{gray!10}Claude Monet & 0.67 & 0.41 & 0.92 & 0.42 & 0.33 \\
Camille Pissarro & 0.80 & 0.86 & 0.92 & 0.87 & \textbf{0.00$_\ast$} \\
\rowcolor{gray!10}Pierre-Auguste Renoir & \textbf{1.00$^\ast$} & \textbf{1.00$^\ast$} & \textbf{1.00} & \textbf{1.00$^\ast$} & 0.47 \\
Vincent van Gogh & \textbf{1.00} & 0.96 & \textbf{1.00} & \textbf{0.98} & 0.10 \\
\hline
\end{NiceTabular}
\end{adjustbox}
\label{tab:impressionists_wasserstein}
\end{table}
We expected to have less significant results, since we are comparing artists with very similar styles: instead, we obtained a perfect classification in this case too, even if they belong to the same artistic current. 
This tells us that, while appearing similar to the unfamiliar viewer, these painters have topologically unique styles. 
In more detail, the hardest to distinguish were Manet, for the Impressionists, and Haldankar, for the Academic Realists: only the bottleneck distance for the blue channel in $1$-PH sets them apart from their group. 
It was curious that the same distance, channel, and dimension worked best for both. 
For both currents, contrary to what happened in the first set of experiments, the $1$-Wasserstein distance was better suited for classification than the bottleneck distance. 
As in the first experiments, van Gogh is consistently closer to all other paintings in the bottleneck distance, but mostly farther away in the $1$-Wasserstein, with the $1$-PH edge channel as the only exception. 
For the Academic Realists, Roy is extremely easy to distinguish, not only with a significant $p$-value in each case, but also almost always achieving either the minimum or the maximum value. 
Mazumdar was harder (but not impossible) to identify with the bottleneck distance, but extremely easy with the $1$-Wasserstein, suggesting that, for the Academic Realists, all features, and not only the longest ones, play a role.
\begin{table}[h]
\caption{The $p$-values for academic realism artists, with the bottleneck distances for all channels, with $0$-PH on the top and $1$-PH at the bottom. Significant $p$-values are in bold; the symbol $^\ast$ (resp. $_\ast$) means the initial sample had the maximum (resp. minimum) average distance.}
\begin{adjustbox}{width=.69\linewidth,center}
\begin{NiceTabular}{l*{5}{c}}
\toprule
\multirow{2}{*}{Artist} &
\multicolumn{5}{c}{$0$-PH Bottleneck} \\
\cmidrule(lr){2-6}
 & Red 
 & Green 
 & Blue
 & Gray
 & Edge \\
\midrule
Sawlaram L. Haldankar & 0.18 & 0.28 & 0.06 & 0.23 & 0.52 \\
\rowcolor{gray!10}Hemendranath Mazumdar & \textbf{0.98} & 0.95 & 0.67 & 0.97 & 0.53 \\
M. F. Pithawala & 0.66 & 0.67 & 0.70 & 0.62 & 0.59 \\
\rowcolor{gray!10}Raja Ravi Varma & 0.34 & 0.26 & 0.74 & \textbf{0.00} & 0.49 \\
Jamini Roy & \textbf{1.00$^\ast$} & \textbf{1.00$^\ast$} & \textbf{1.00$^\ast$} & \textbf{1.00$^\ast$} & \textbf{1.00$^\ast$} \\
\hline
\end{NiceTabular}
\end{adjustbox}
\begin{adjustbox}{width=.69\linewidth,center}
\begin{NiceTabular}{l*{5}{c}}
\toprule
\multirow{2}{*}{Artist} &
\multicolumn{5}{c}{$1$-PH Bottleneck} \\
\cmidrule(lr){2-6}
 & Red 
 & Green 
 & Blue
 & Gray
 & Edge \\
\midrule
Sawlaram L. Haldankar & 0.10 & 0.04 & \textbf{0.01} & 0.03 & 0.11 \\
\rowcolor{gray!10}Hemendranath Mazumdar & 0.92 & 0.73 & 0.73 & 0.94 & 0.94 \\
M. F. Pithawala & \textbf{1.00$^\ast$} & 0.96 & 0.96 & \textbf{1.00} & \textbf{1.00$^\ast$} \\
\rowcolor{gray!10}Raja Ravi Varma & \textbf{0.99} & 0.62 & 0.94 & 0.83 & \textbf{0.99} \\
Jamini Roy & \textbf{1.00$^\ast$} & \textbf{1.00$^\ast$} & \textbf{1.00$^\ast$} & \textbf{1.00$^\ast$} & \textbf{1.00} \\
\hline
\end{NiceTabular}
\end{adjustbox}
\label{tab:ac_bottleneck}
\end{table}
\begin{table}
\caption{The $p$-values for academic realism artists, with the $1$-Wasserstein distances for all channels, with $0$-PH on the top and $1$-PH at the bottom. Significant $p$-values are in bold; the symbol $^\ast$ (resp. $_\ast$) means the initial sample had the maximum (resp. minimum) average distance.}
\begin{adjustbox}{width=.69\linewidth,center}
\begin{NiceTabular}{l*{5}{c}}
\toprule
\multirow{2}{*}{Artist} &
\multicolumn{5}{c}{$0$-PH $1$-Wasserstein} \\
\cmidrule(lr){2-6}
 & Red 
 & Green 
 & Blue
 & Gray
 & Edge \\
\midrule
Sawlaram L. Haldankar & 0.35 & 0.43 & 0.51 & 0.43 & 0.49 \\
\rowcolor{gray!10}Hemendranath Mazumdar & \textbf{0.00$_\ast$} & \textbf{0.00$_\ast$} & \textbf{0.00$_\ast$} & \textbf{0.00$_\ast$} & \textbf{0.00$_\ast$} \\
M. F. Pithawala & \textbf{0.00$_\ast$} & \textbf{0.00$_\ast$} & \textbf{0.00$_\ast$} & \textbf{0.00$_\ast$} & \textbf{0.00$_\ast$} \\
\rowcolor{gray!10}Raja Ravi Varma & \textbf{0.00$_\ast$} & \textbf{0.00$_\ast$} & \textbf{0.00$_\ast$} & \textbf{0.00$_\ast$} & 0.94 \\
Jamini Roy & \textbf{1.00$^\ast$} & \textbf{1.00$^\ast$} & \textbf{1.00$^\ast$} & \textbf{1.00$^\ast$} & \textbf{0.00$_\ast$} \\
\hline
\end{NiceTabular}
\end{adjustbox}
\begin{adjustbox}{width=.69\linewidth,center}
\begin{NiceTabular}{l*{5}{c}}
\toprule
\multirow{2}{*}{Artist} &
\multicolumn{5}{c}{$1$-PH $1$-Wasserstein} \\
\cmidrule(lr){2-6}
 & Red 
 & Green 
 & Blue
 & Gray
 & Edge \\
\midrule
Sawlaram L. Haldankar & 0.35 & 0.51 & 0.55 & 0.49 & 0.30 \\
\rowcolor{gray!10}Hemendranath Mazumdar & \textbf{0.00$_\ast$} & \textbf{0.00$_\ast$} & \textbf{0.00$_\ast$} & \textbf{0.00$_\ast$} & \textbf{0.00$_\ast$} \\
M. F. Pithawala & 0.60 & 0.25 & \textbf{0.00$_\ast$} & 0.39 & 0.12 \\
\rowcolor{gray!10}Raja Ravi Varma & 0.14 & 0.00 & 0.03 & \textbf{0.00$_\ast$} & 1.00 \\
Jamini Roy & \textbf{1.00$^\ast$} & \textbf{1.00$^\ast$} & \textbf{1.00$^\ast$} & \textbf{1.00$^\ast$} & \textbf{1.00$^\ast$} \\
\hline
\end{NiceTabular}
\end{adjustbox}
\label{tab:ac_wasserstein}
\end{table}

\begin{figure}[!h]
\centering
\begin{subfigure}[t]{.50\linewidth}
\centering
\includegraphics[scale=.335]{apple.png}
\label{subfig_original_apple_forai}
\end{subfigure}
\begin{subfigure}[t]{.48\linewidth}
\centering
\includegraphics[scale=.09]{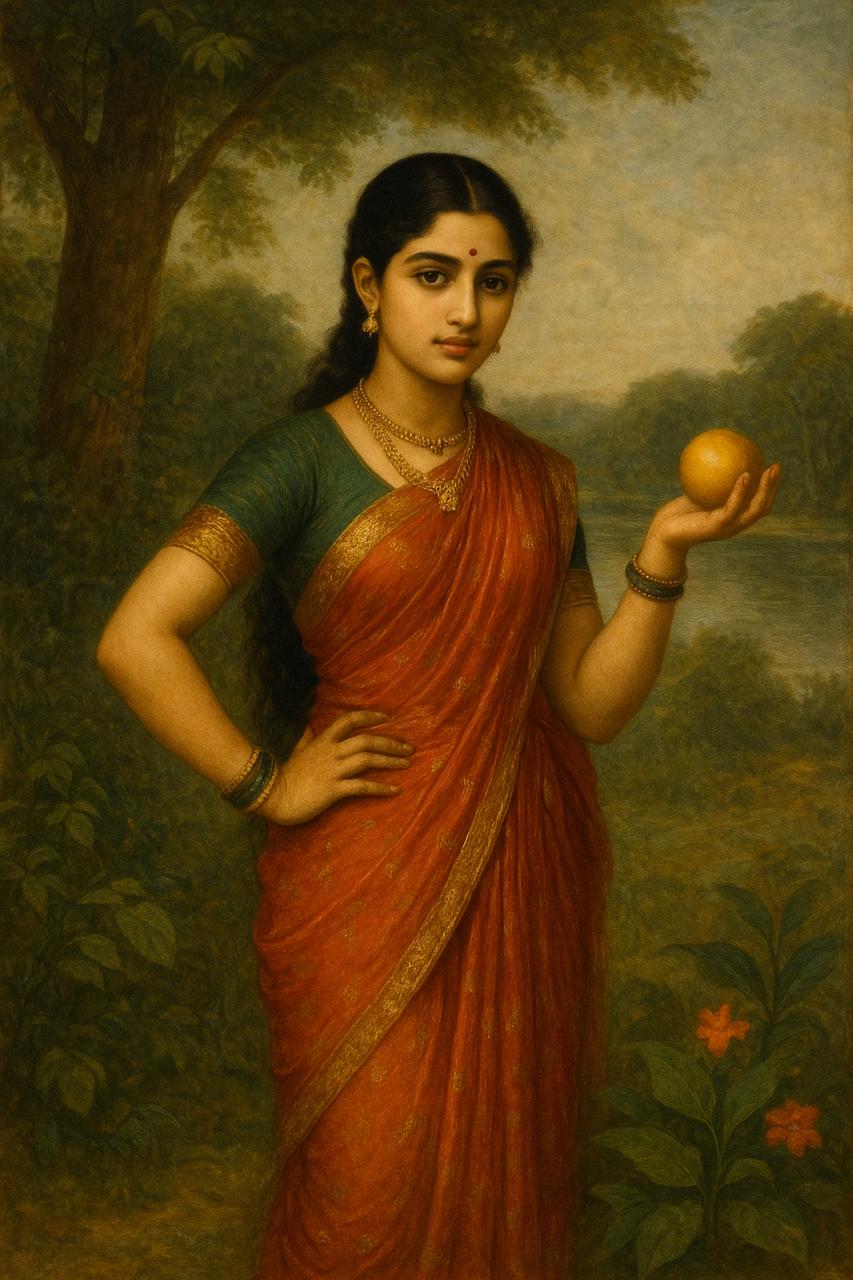}
\label{subfig_ai_apple}
\end{subfigure}
\caption{Original painting by Raja Ravi Varma (left) and an AI-generated image in Raja's style (right).}
\label{fig_apple_w_ai}
\end{figure}

The third and last set of experiments answers question \hyperlink{(iii)}{(iii)}. 
We performed, in this case, only one permutation test per channel, namely putting the 10 paintings of an artist (either Klimt or Raja) in sample $A$ and the AI-generated image in sample $B$. 
Since we had only 11 paintings in each set, we could not perform a conclusive permutation test. 
Indeed, in this case, there are only 11 possible permutations, and they each occur with $\frac{1}{11}$ frequency, which cannot be in the 5\% percentile. 
Therefore, we only recorded whether the initial average distance was the maximum, the minimum, or neither.
The outputs are reported in \cref{tab:ai_klimt} for the Klimt and \cref{tab:ai_raja} for Raja. 
Even if, for this limited data, we cannot confirm our findings with statistical certainty, we have strong indications that PH still captures the difference between AI-generated images and original paintings. 
In particular, the blue channel is a strong indicator, both for Klimt and Raja, of whether the painting is original or not. 
The result is particularly interesting since, to the naked eye, the difference between the paintings is not as striking, as shown in \cref{fig_apple_w_ai}: the subject and the color palette look very similar.
\begin{table}
\caption{Results of whether the average distances from the AI-generated image in Klimt style to all other images were the min, the max, or neither (denoted by -) of all the average distances obtained by permutations for all five channels.}
\begin{adjustbox}{width=.69\linewidth,center}
\begin{NiceTabular}{l*{5}{c}}
\toprule
\multirow{2}{*}{\makecell{Distance and  \\ homological dimension}} & 
\multicolumn{5}{c}{Klimt vs AI Klimt} \\ 
\cmidrule(lr){2-6}
 & Red 
 & Green 
 & Blue
 & Gray
 & Edge \\
\midrule
$0$-PH Bottleneck & min & - & Max & - & - \\
\rowcolor{gray!10}$1$-PH Bottleneck & - & - & Max & - & - \\
$0$-PH $1$-Wasserstein & - & - & Max & - & - \\
\rowcolor{gray!10}$1$-PH $1$-Wasserstein & - & Max & Max & Max & - \\
\hline
\end{NiceTabular}
\end{adjustbox}
\label{tab:ai_klimt}
\end{table}
\begin{table}[!h]
\caption{Results of whether the average distances from the AI-generated image in Raja style to all other images were the min, the max, or neither (denoted by -) of all the average distances obtained by permutations for all five channels.}
\begin{adjustbox}{width=.69\linewidth,center}
\begin{NiceTabular}{l*{5}{c}}
\toprule
\multirow{2}{*}{\makecell{Distance and  \\ homological dimension}} & 
\multicolumn{5}{c}{Raja vs AI Raja} \\ 
\cmidrule(lr){2-6}
 & Red 
 & Green 
 & Blue
 & Gray
 & Edge \\
\midrule
$0$-PH Bottleneck & - & - & - & - & - \\
\rowcolor{gray!10}$1$-PH Bottleneck & - & - & Max & - & - \\
$0$-PH $1$-Wasserstein & - & - & Max & - & - \\
\rowcolor{gray!10}$1$-PH $1$-Wasserstein & - & - & - & - & - \\
\hline
\end{NiceTabular}
\end{adjustbox}
\label{tab:ai_raja}
\end{table}

\section{Discussion and conclusion}\label{sec:discussion}

Our results show that persistent homology is able to capture stylistic differences between artists, both inside and outside the same artistic current. 
While not all channels are always significant, for every artist there is always at least one that can identify them, either in the bottleneck or in the $1$-Wasserstein. 
What was most remarkable for the authors is that the information carried by the \textit{non}-significant features is as meaningful as the information carried by the significant ones, if not more. 
Indeed, a significant feature tells us that the artist has a ``peculiar'' way of handling that feature, while a non-significant $p$-value tells us that that feature is ``common'' among several artists. 
We can thus pick up not only the differences but also the similarities between artists. 
\newline
We remark that, in the above paragraph, the words ``peculiar'' and ``common'' are used from a mathematical perspective and not from a traditional artistic sense: our backgrounds are in mathematics and data science, not in art and art history. 
We invite experts in these fields to investigate our findings further in their language.
\newline
The only non-negligible drawback of using persistent homology for analyzing style is the computational cost: while computing the $0$- and $1$-PH of a single image can be done quickly (i.e., less than a second), performing all the needed permutation tests (and all the distance computations they entail) can take several minutes if not hours. 
For example, computing all the bottleneck distances for the first set of experiments took one hour, and the permutation tests on those distances took two and a half hours on a 2024 Lenovo Yoga laptop with 16 processors.  
However, as we were not focusing on runtime, the code is not optimized, and the machine was performing other (minor) tasks while the experiments were running, so possibly the exact runtime is lower (even if we do not expect it to be so by much). 
Thus, the current implementation can certainly be improved in runtime, but we doubt it can be brought down to a few minutes.
Therefore, to be analyzed with this method, a collection should contain fewer than 500 images to be manageable on a common laptop. 
Nevertheless, with the insights provided in this paper, one can compute the persistent homology of the various channels of a painting and gain an understanding of the style of the artists by observing its topological features. 
\newline
The authors would be particularly interested in seeing this work replicated and expanded by groups that include art experts, to provide better insights on the artists and their styles, using images at different, higher resolutions. 
Indeed, it would be interesting to see what PH can tell us not only, as in this paper, about the general composition of a painting, but also more precisely about the brush strokes and color application. 
This could be done, for example, by analyzing high-resolution images of paintings.
Moreover, in this work, we proved via permutation tests that PH has strong classification powers, but PH has more capability than more common data analysis methods and can be more than a classifier: it is a feature extractor, which means that, using software like \cite{oat}, it is possible to pinpoint in the painting where the PH intervals come from, and thus obtain finer insights about which topological feature is contributing to a given artistic style. 
\newline
As remarked above, in this work, we selected some specific paintings and painters, but the analysis is independent of this choice. 
Therefore, the method outlined here can be used on virtually any image. 
Moreover, since cubical complexes can also be computed for 3D objects in much the same way, the analysis we presented can be equally applied to statues or other 3D objects for which a complete scan is available.

\section*{Declaration of Generative AI Use}

During the preparation of this work, the authors used ChatGPT to generate some of the images to be used in the experiments, with the precise goal of classifying original from AI-generated images. 
The authors have reviewed and edited the content and take full responsibility for its integrity.

\section*{Acknowledgement(s)}
The authors thank H\aa vard Bakke Bjerkevik for feedback on the writing of the first draft of this work. 
This paper is currently under consideration at Pattern Recognition Letters.

\section*{Disclosure statement}

The authors declare no conflicts of interest.

\section*{Funding}

This work was funded in part by a grant from the Simons Foundation [MPS-TSM-00007525, BG].

\end{document}